\documentclass[letterpaper, 10 pt, conference]{ieeeconf} 
\IEEEoverridecommandlockouts
                                      \usepackage{cite}
\usepackage{amsmath,amssymb,amsfonts}
\usepackage{algorithmic}
\usepackage{graphicx}
\usepackage{textcomp}
\usepackage{xcolor}
\def\BibTeX{{\rm B\kern-.05em{\sc i\kern-.025em b}\kern-.08em
    T\kern-.1667em\lower.7ex\hbox{E}\kern-.125emX}}

\usepackage{algorithm}

\usepackage{balance} 

\usepackage{mathtools}

\usepackage{multirow} \usepackage{hhline}

\graphicspath{{figures/}} 

\usepackage{subfigure}

\usepackage{array}

\usepackage{array}

\begin{document}

\title{\LARGE \bf
Integrable Whole-body Orientation Coordinates for Legged Robots
}

\author{Yu-Ming Chen$^{1,3}$, Gabriel Nelson$^{2,3}$, Robert Griffin$^{4}$, Michael Posa$^{1}$ and Jerry Pratt$^{3,4}$   \thanks{$^{1}$The authors are with the General Robotics, Automation, Sensing and Perception (GRASP) Laboratory, University of Pennsylvania, Philadelphia, PA 19104, USA
        {\tt\small \{yminchen, posa\}@seas.upenn.edu}}\thanks{$^{2}$The author is with the Boston Dynamics Artificial Intelligence Institute, 145 Broadway, Cambridge, MA 02142
        {\tt\small gnelson@theaiinstitute.com}}\thanks{$^{3}$The authors are with Boardwalk Robotics, 417 E Zaragoza St, Pensacola, FL 32502
        {\tt\small jerry.pratt@boardwalkrobotics.com}}\thanks{$^{4}$The authors are with the Florida Institute for Human and Machine Cognition (IHMC), 40 S Alcaniz St, Pensacola, FL 32502, USA
        {\tt\small rgriffin@ihmc.org}}}

\maketitle

\begin{abstract}

Complex multibody legged robots can have complex rotational control challenges.
In this paper, we propose a concise way to understand and formulate a \emph{whole-body orientation} that
(i) depends on system configuration only and not a history of motion,
(ii) can be representative of the orientation of the entire system while not being attached to any specific link, and
(iii) has a rate of change that approximates total system angular momentum.
We relate this orientation coordinate to past work, and discuss and demonstrate, including on hardware, several different uses for it.

\end{abstract}

\section{Introduction}

Many legged robots are best represented by nontrivial multibody dynamic models.  For these systems, reduced-order coordinates have been widely used for model-based planning and control, since these low-dimensional coordinates can capture the bulk of the robot's dynamics 
\cite{wensing2022optimization, Kuindersma14, kajita1991study, chen2020optimal}.
These models and their coordinates are often derived from physical intuition \cite{kajita1991study} or computed via optimization \cite{chen2020optimal}.
The total system center of mass (CoM) is likely the most well-known of these model-based coordinates. 
It responds to the total net forces acting on the robot according to Newton's 2nd Law, and gives us an overall translational location for the robot, providing extremely useful information for controlling locomotion.

When we look more closely at complex tasks, such as human-like walking and running, back-flips or aggressive turning, we typically need to also consider the orientation of the robot.
In regulating orientation, many researchers have used models based on centroidal angular momentum 
\cite{orin2008centroidal, Dai14}.
Centroidal angular momentum models are usually based on velocity level constraints, and tend not to readily produce an (unique) absolute orientation coordinate for the entire system.
Other researchers use a single rigid body (SRB) model \cite{bledt2018cheetah}, where a single SO(3) coordinate represents the entire robot's orientation.
The choice for this single body may stem from the morphology and mass distribution of a specific robot.
For example, for robots with a heavy torso and light limbs, such as the MIT Mini Cheetah \cite{katz2019mini}, the torso orientation can act as a good proxy for total system orientation. 
However, for robots with relatively heavy and/or long limbs, where appreciable mass is distributed throughout the system and far from the CoM (e.g IHMC Nadia \cite{NadiaHumanoid} and Agility Robotics Cassie \cite{batke2022optimizing}), the coordinate choice for system orientation is much less clear.
Proposing a useful whole-body orientation coordinate for these types of systems, that is not attached to any specific link on the robot, is the focus of this paper.

Unlike the CoM, a weighted averaging of each link's orientation does not produce a consistently meaningful whole-body orientation.
To derive this orientation, some have used angular excursion, which is an integral of an angular velocity about the CoM derived from centroidal angular momentum \cite{zordan2014control}.
However, there is not a unique angular excursion value for a defined joint configuration, as the integral is path dependent.
Another approach uses the principal axes of the whole-body inertia tensor to derive a whole-body orientation \cite{du2021meaningful}.
This approach also relies on the history of the axes in order to produce a meaningful continuous orientation.

\begin{figure}
 \centering
 \includegraphics[width=1\linewidth]{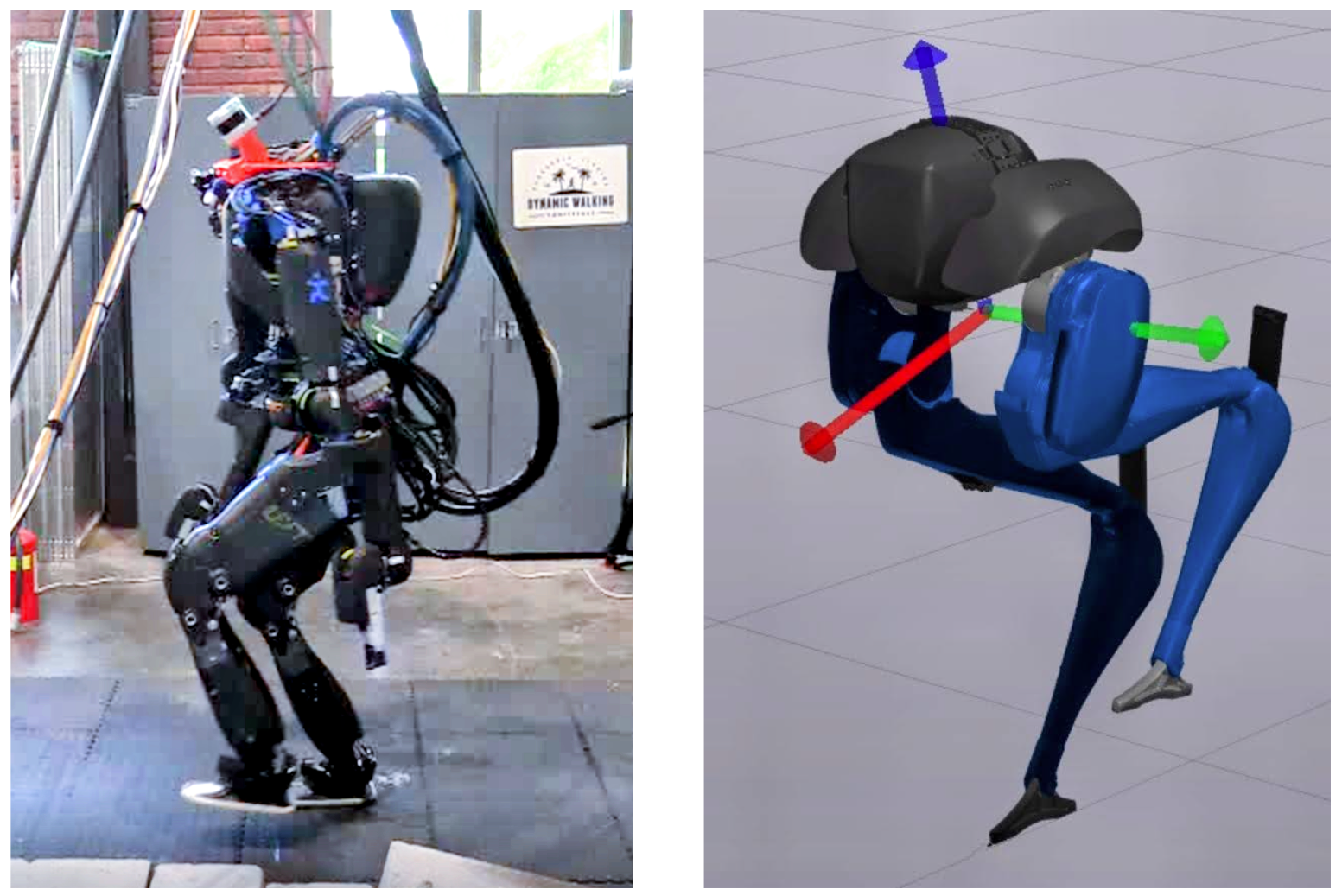}
 \caption{Left: The humanoid Nadia walking with arm swing and spine yaw rotation induced by tracking WBO. Right: The simulated biped Cassie running and turning with the whole-body orientation (WBO) coordinate visualized at the CoM.
}
 \label{fig:hardware_side_by_side_comparison}
\end{figure}

Researchers in geometric mechanics have, for about a decade, devised ways of finding history-independent optimal coordinates, called the \emph{minimum perturbation coordinates} \cite{travers2013minimum, hatton2011geometric, hatton2015nonconservativity}.
However, to our knowledge, no literature has yet shown the application of these coordinates to a complex dynamic robot (except for Boston Dynamics' patent \cite{khripin2016natural}), and they remain less prevalent than other approaches. 

Similar to \cite{travers2013minimum}, our focus is to create an \emph{integrable} (history and path invariant) measure of whole-body orientation for multi-link humanoid and legged robots, where this orientation coordinate (e.g. an angle, or Euler angles, or quaternion, etc.) has dynamically analogous behavior to a CoM, but crucially in an angular sense.
For example, once formulated, our desire would be that, should there be no external moments acting on the system, this orientation coordinate would remain at rest or rotate at a constant speed.
We call this coordinate the integrable whole-body orientation (abbreviated WBO in this paper), though it has been called the \emph{minimum perturbation coordinates} for general coordinates (other than SO(3) coordinates) \cite{hatton2011geometric, travers2013minimum}.
As such, we design the WBO to have the same mathematical form as other forward kinematics quantities
such as a hand position, or the CoM.
Thus, most or all tools and techniques that apply to forward kinematics can be applied to a WBO, such as inverse kinematics, task-space control and planning, etc.

\subsection*{Contributions of this paper}

\begin{enumerate}
\item Presenting a simple example that clarifies the concept behind the WBO, with clear definitions of how the problem and solution are structured.
\item Providing a concise algorithm that shows how to find an WBO representation for a general multibody robot in 3D. 
\item Demonstrating the use of the WBO on hardware (the humanoid robot Nadia) and in simulation (the biped robot Cassie); showing improvements in reducing angular momentum oscillation and foot yaw moment.
\end{enumerate}

The paper is organized as follows.
Section \ref{sec:formulation} uses simple 2D examples to introduce our WBO.
Section \ref{sec:acom_optimization} extends this to complex 3D systems and demonstrates the WBO algorithm with Nadia and Cassie.
Sections \ref{sec:walking_example} and \ref{sec:running_example} apply the derived WBO in walking and running controllers.
Section \ref{sec:conclusion} summarizes this paper and describes future work.

\begin{figure}
 \centering
 \includegraphics[width=0.8\linewidth]{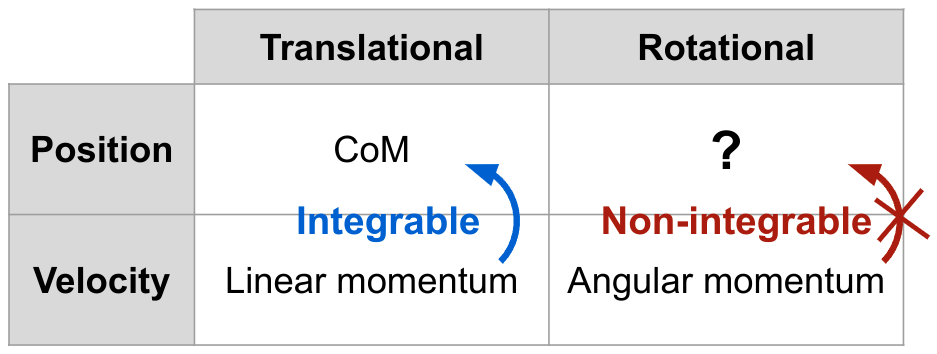}
 \caption{
 While linear momentum is integrable, angular momentum is \emph{generally} not \cite{nakamura1990nonholonomic, saccon2017centroidal}.
}
 \label{fig:acom_table}
\end{figure}

\section{WBO of simple systems}\label{sec:formulation}

\subsection{Motivation and Problem Definition} \label{sec:def_and_motivation}

Fig. \ref{fig:acom_table} conceptually compares standard centroidal translational and rotational quantities often used in whole-body control.
While the total system CoM (a position), and linear and angular momenta are concrete properties of a multibody system, there is in general no unique rotational coordinate (an orientation) corresponding to system CoM.
Differentiation of system CoM (scaled by total mass) will arrive at linear momentum.
Angular momentum though, for general systems, is \textit{not} integrable \cite{nakamura1990nonholonomic}, highlighting that angular momentum does not represent differential motion along any unique history or path invariant manifold in configuration space.

Thus, as discussed above, the \emph{WBO} is an effort to approximate an angular measurement for the upper-right quadrant in Fig.  \ref{fig:acom_table}.
For general systems, it will be an angle, a set of Euler angles, or a quaternion, etc.
Its value is a manufactured quantity (but not without physical relevance).
It is the result of a design process that can take various forms, and we present one approach in this paper that we have found useful.

Often this topic involves a larger, typically \emph{formal}, mathematical discussion about differential calculus and geometric mechanics, which we will consider beyond the scope of this paper.
Deeper treatments can be found in \cite{travers2013minimum, hatton2011geometric, hatton2015nonconservativity}.
Our goal is rather to provide a concise WBO formulation recipe, and demonstrate its initial use on a few legged robots.

We propose the following benefits from using a WBO for legged robots:
(1) As a dynamically relevant WBO for controller tracking: e.g. providing a feedback signal for a proportional term on WBO control;
(2) For planning WBO motions or changes;
(3) For encouraging low angular momentum behavior for steady-state walking or running \cite{popovic2005ground, erez2012trajectory, miyata2019walking}.
We believe that a suitable WBO measure for an anthropomorphic robot can aid in producing more natural looking movement, since the whole-body orientation control can be achieved by regulating the WBO directly, rather than controlling the orientation of some specific \emph{base} link (often the pelvis or torso of the robot).
This means the \emph{base} link is now free to be treated as just another link on the robot, available for other user-specified objectives:
e.g. smoothing system CoM motion by extending leg reach or stride length, stepping while ascending/descending terrain, whole-body reaching motions, etc.

\subsection{The Bar-and-Flywheel Model} \label{sec:bar_and_flywheel}

We propose the following simple example as an aid in understanding the WBO that we intend to find.
The example has simple and complex versions, which are meant to demonstrate what the WBO does and does not represent.

\begin{figure}[!t]
\centering
\subfigure[The starting configuration ($\phi = 0$)]{
\includegraphics[width=0.49\textwidth]{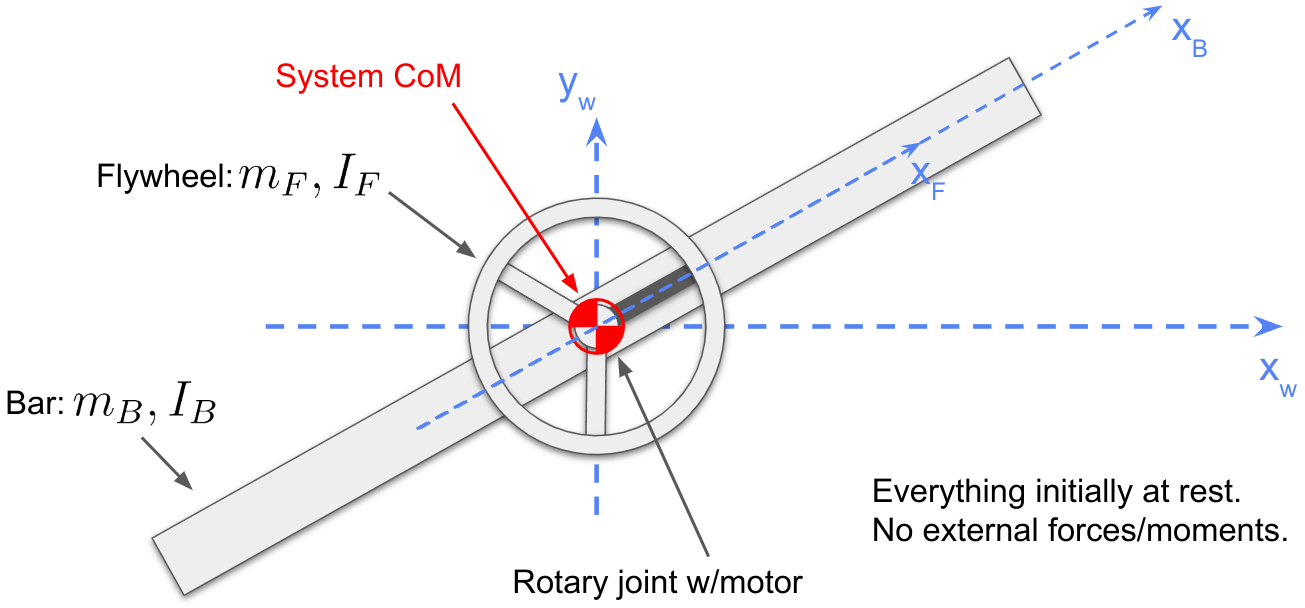}
\label{fig:bar_and_flywheel_1}}
\subfigure[The ending configuration ($\phi>0$)]{
\includegraphics[width=0.49\textwidth]{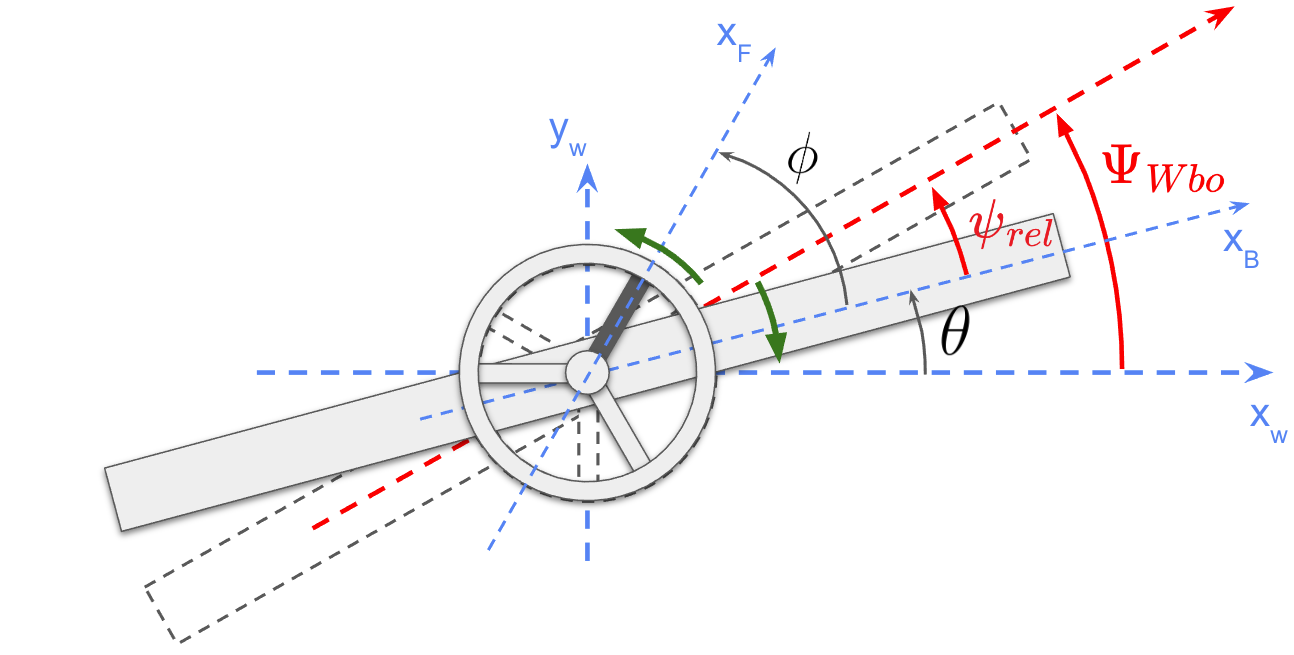}
\label{fig:bar_and_flywheel_4}}
\caption{Bar-and-Flywheel model (an integrable system). The motor rotates the flywheel counter-clockwise.}
\label{fig:bar_and_flywheel}
\end{figure}

Fig. \ref{fig:bar_and_flywheel} shows a planar ``Bar-and-Flywheel” model:
a long solid bar attached, at its CoM, to the axis of a flywheel via a rotary joint.
The bar and flywheel have mass properties as indicated, and are free floating with no gravity or external forces acting.
$\theta$ is the bar orientation in an inertially fixed world-frame, and $\phi$ is the flywheel orientation relative to the bar.
A motor actuates the joint between the bar and flywheel.
Thus, the bar represents the \emph{base} of a multibody system, and the flywheel is an outboard body connected to the base by an actuated joint.
The angular momentum about the CoM, also called the \emph{centroidal angular momentum} (CAM), is
\vspace{-0.5mm} 
\begin{equation} \label{eq:BF_H_com}
  H_{CoM} = (I_B + I_F) \dot{\theta} + I_F \dot{\phi}.
\end{equation}
In our definitions, we will delineate \emph{base} orientation ($\theta$ in this example; in general a SO(3) representation) from \emph{joint} positions or \emph{joint} configuration ($\phi$ in this example; in general represented by the vector $q$).

Let the system start at rest with $\phi = 0$ (Fig. \ref{fig:bar_and_flywheel_1}).
If the motor drives the flywheel counter-clockwise, the reaction torque will rotate the bar in the opposite direction. 
These rotation directions are indicated by the green arrows.
Fig. \ref{fig:bar_and_flywheel_4} shows both the starting (dashed lines) and ending configurations.
We would like to understand how \emph{the bar} moves due to the motion of the flywheel.

An important relationship often used in these problems is the \emph{reconstruction equation}, which includes a \emph{local connection} \cite{hatton2011geometric}.
The reconstruction equation describes what we have simulated going from Fig. \ref{fig:bar_and_flywheel_1} to Fig. \ref{fig:bar_and_flywheel_4}:
It maps velocities in joint space ($\dot{\phi}$) to the velocity of the base ($\dot{\theta}$), as the base will counter-rotate due to changes in joint space.
In this example ($H_{CoM} = 0$), this reconstruction equation is
\vspace{-2mm} 
\begin{equation} \label{eq:BF_Recon}
  \dot{\theta} = -\frac{I_F}{I_B + I_F} \dot{\phi},
\vspace{-1mm} 
\end{equation}
\noindent where the coefficient in front of $\dot{\phi}$ (without the negative sign) is the local connection.
We can see that Eq. \eqref{eq:BF_Recon} can be integrated directly and also lets us predict the change in $\theta$. 
Let $\Delta \theta$ and $\Delta \phi$ be the changes in the angles. 
We define
\vspace{-2mm} 
\begin{equation} \label{eq:BF_ACoMrel}
  \psi_{rel} \triangleq -\Delta\theta = \frac{I_F}{I_B + I_F} \Delta \phi,
\end{equation}
and we note that $\Delta \phi = \phi$ since the starting position of $\phi$ is $0$.
In Fig. \ref{fig:bar_and_flywheel_4}, we label these various angles. 
The \emph{initial} orientation of the bar, relative to the world, will be called $\Psi_{Wbo}$.
In a general configuration, we can predict the final orientation of the system using Eq. \eqref{eq:BF_ACoMrel}:
\vspace{-2mm} 
\begin{equation} \label{eq:BF_ACoM}
  \Psi_{Wbo} = \theta + \psi_{rel} = \theta + \frac{I_F}{I_B + I_F} \phi.
\end{equation}
This $\Psi_{Wbo}$ is the WBO of the system.
Note that $\psi_{rel}$ is the \emph{relative} orientation of the WBO to the base (bar), such that the final $\Psi_{Wbo}$ will be a base orientation relative to the world plus $\psi_{rel}$.

For this simple system, conservation of angular momentum dictates that $\Psi_{Wbo}$ will never change, regardless of how we actuate the motor.
Differentiating Eq. \eqref{eq:BF_ACoM}, we have (after some rearrangement)
\vspace{-1mm} 
\begin{equation} \label{eq:BF_ACoMdot}
  (I_B + I_F) \dot{\Psi}_{Wbo} = (I_B + I_F) \dot{\theta} + I_F \dot{\phi}.
\end{equation}
Note that the right hand sides of Eq. \eqref{eq:BF_H_com} and Eq. \eqref{eq:BF_ACoMdot} are the same. Thus:
\vspace{-2mm} 
\begin{equation} \label{eq:BF_HCoM-ACoMdot}
  H_{CoM} = (I_B + I_F) \dot{\Psi}_{Wbo}.
\end{equation}
Eq. \eqref{eq:BF_HCoM-ACoMdot} shows that the WBO has behavior analogous to a center of mass, \emph{but in a rotational sense}: it represents an underlying orientation state for the system that can only be changed by external moments.
The effective WBO mass moment of inertia (MoI) is, not surprisingly, the sum of the MoI's of the bar and flywheel.

The $H_{CoM}$ above is an integrable differential form, 
meaning it can be derived from the differentiation of a manifold in configuration space, independent of the joints' history. 
This manifold can be expressed exactly as Eq. \eqref{eq:BF_ACoM}.

Also, looking at the left and right-hand sides of Eq. \eqref{eq:BF_HCoM-ACoMdot}, we could think of a system having two representations of angular momentum: 
the \emph{actual} system angular momentum about the CoM in Eq. \eqref{eq:BF_H_com}, 
and an \emph{approximated} angular momentum based on $\dot{\Psi}_{Wbo}$  in Eq. \eqref{eq:BF_HCoM-ACoMdot}.
For this integrable example system, we see that these angular momentum representations are equivalent, though for general systems (usually non-integrable) they are not.

\begin{figure}
 \centering
 \includegraphics[width=0.9\linewidth]{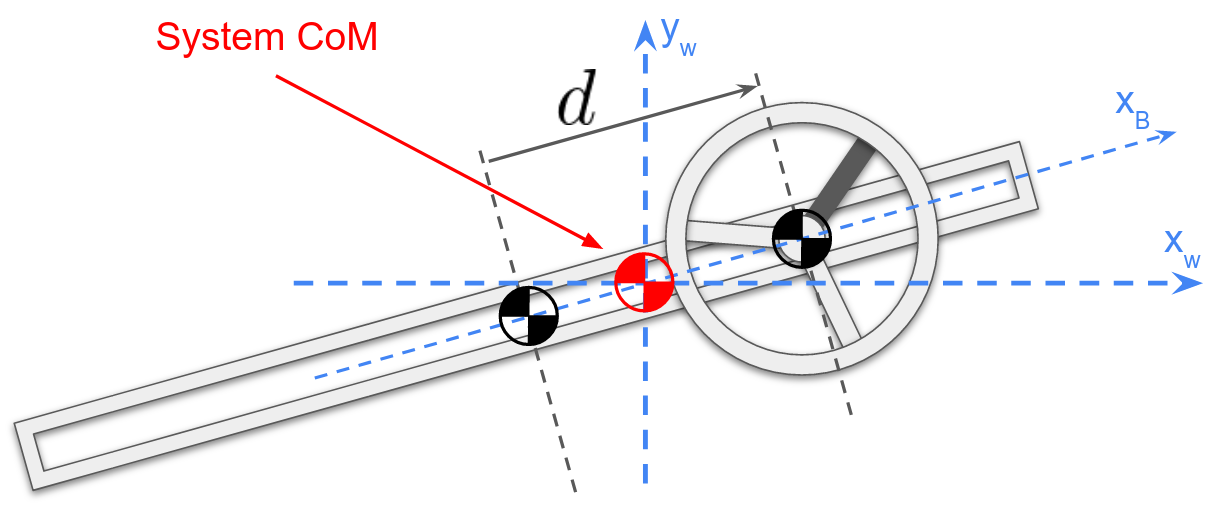}
 \caption{Bar-and-Flywheel model with an actuated prismatic joint (a non-integrable system).
  This system can reorient itself without external torques.
 For example, the entire system can rotate counter-clockwise if the joints $(d, \phi)$ repeat the following cyclic motion: $(0, 0) \rightarrow (l/2, 0) \rightarrow (l/2, \pi/2) \rightarrow (0, \pi/2) \rightarrow (0, 0)$, where $l$ is the length of the bar.
}
 \label{fig:bar_and_flywheel_with_slot}
\end{figure}

We now extend this model by adding a long slot that allows the flywheel to slide along the length of the bar (Fig. \ref{fig:bar_and_flywheel_with_slot}).
A linear actuator controls this movement, and $d$ is the offset of the flywheel axis from the CoM of the bar.
This change increases the size of our joint space ($q$) to 2 dimensions: $q = [d, \phi]^T$.
The new CAM is
\begin{equation} \label{eq:BF_H_com_slot}
  H_{CoM} = (I_B + I_F + \frac{m_B m_F}{m_B + m_F} d^2) \dot{\theta} + I_F \dot{\phi}.
\end{equation}
If $d$ is fixed, then this system is much the same as the simpler Bar-and-Flywheel model above.
If, though, $d$ is allowed to change, the angular momentum becomes non-integrable, and
the goal of our WBO formulation becomes \emph{contriving a differentiable and time independent function for $\psi_{rel}$ that, when differentiated with respect to $q$, maximally approximates the local connection of the actual system over a user-defined region of joint-space.}
We will call this contrived function $\tilde{\psi}_{rel}(q)$.
Like Eq. \eqref{eq:BF_Recon}, the reconstruction equation is found by setting $H_{CoM}=0$ in Eq. \eqref{eq:BF_H_com_slot} and solving for $\dot{\theta}$:
\begin{equation} \label{eq:BF_cnx1}
  \dot{\theta} = -
\begin{bmatrix} 0 \;\; \frac{I_F}{I_B + I_F + \frac{m_B m_F}{m_B + m_F} d^2} \end{bmatrix}_{1 \times 2} 
  \begin{bmatrix} \dot{d} \\ \dot{\phi} \end{bmatrix}_{2 \times 1} ,
\end{equation}
where the 1x2 matrix is the local connection, which is a function of $d$.  
In mathematical terms, the goal (stated above) is finding a differentiable function $\tilde{\psi}_{rel}(q)$, such that
\begin{equation} \label{eq:BF_final}
  \begin{bmatrix} \frac{\partial{\tilde{\psi}_{rel}(q)}}{\partial{d}} \; \frac{\partial{\tilde{\psi}_{rel}(q)}}{\partial{\phi}} \end{bmatrix}
  \approx
  \begin{bmatrix} 0 \;\; \frac{I_F}{I_B + I_F + \frac{m_B m_F}{m_B + m_F} d^2} \end{bmatrix}.
\end{equation}
Coming up with this function is the core design process, 
bearing in mind that an exact fit is impossible owing to the non-integrability of the physical system.
For instance, an example function for $\tilde{\psi}_{rel}(q)$ could be $c_1\,d^2 \phi + c_2\,d^2 \phi^3 + c_3\,\phi + c_4\,\phi^3$, where the coefficients $c_i$ are found numerically in order to maximize the approximation implied by Eq. \eqref{eq:BF_final} over a user-defined region in joint-space.
Our final WBO now becomes (like Eq. \eqref{eq:BF_ACoM}):
\begin{equation} \label{eq:BF_ACoM_tilde}
  \tilde{\Psi}_{Wbo} \triangleq \theta + \tilde{\psi}_{rel}(q) .
\end{equation}
Finally we note that, mirroring Eq. \eqref{eq:BF_HCoM-ACoMdot}, we now have \emph{actual} and \emph{approximated} representations for angular momentum (approximation resulting from Eq. \eqref{eq:BF_final}):
\begin{equation} \label{eq:BF_HCoM_approx1}
  H_{CoM} \approx \tilde{H}_{CoM}
\end{equation}
with
\begin{equation} \label{eq:BF_HCoM_approx2}
  \tilde{H}_{CoM} = (I_B + I_F + \frac{m_B m_F}{m_B + m_F} d^2) \dot{\tilde{\Psi}}_{Wbo},
\end{equation}
where
\begin{equation} \label{eq:BF_HCoM_approx2}
\dot{\tilde{\Psi}}_{Wbo} = \dot{\theta} +
  \begin{bmatrix} \frac{\partial{\tilde{\psi}_{rel}(q)}}{\partial{d}} \; \frac{\partial{\tilde{\psi}_{rel}(q)}}{\partial{\phi}} \end{bmatrix}
  \begin{bmatrix} \dot{d} \\ \dot{\phi} \end{bmatrix}.
\end{equation}

The main compromise in our approach is approximating a non-integrable differential system with an integrable one.
This ``collapses” explicit representation of the nonholonomic motion of the actual physical system \cite{nakamura1990nonholonomic, hatton2011geometric, hatton2015nonconservativity}.
Our WBO will still measure nonholonomic motions, but its resulting dynamics will not correspond with fidelity to the actual externally applied moments.

\vspace{2mm}

\section{WBO of complex robots}\label{sec:acom_optimization}

\subsection{Extending WBO to 3D} \label{sec:ACoM_3D}

\begin{figure}
 \centering
 \includegraphics[width=0.95\linewidth]{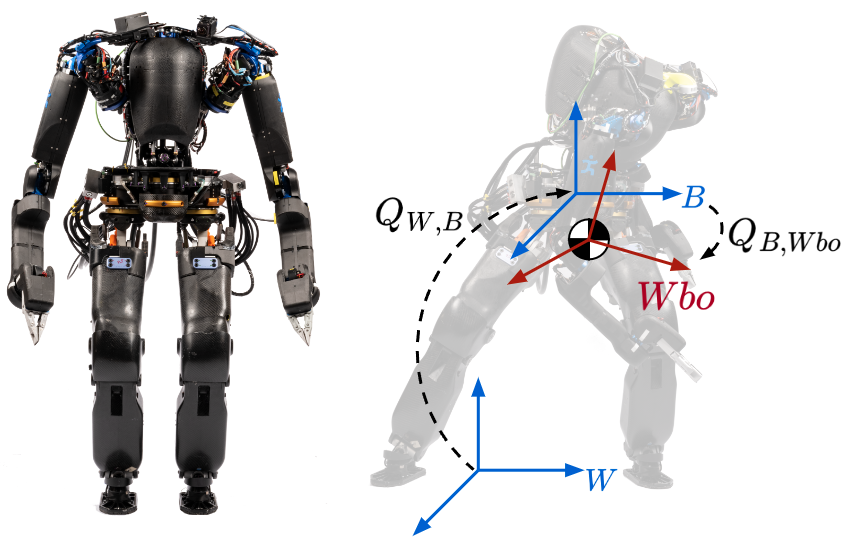}
 \caption{The humanoid Nadia (left) 
 and frames of interest (right).
 $W$, $B$ and $Wbo$ are the world frame, base frame and the WBO frame, respectively.
 $Q_{W,B}$ is the base orientation in the world, and $Q_{B,Wbo}$ is the orientation of the WBO frame relative to the base.
 Translationally, we locate the WBO frame at the CoM for convenience.
\vspace{2mm}
}
 \label{fig:frames_def}
\end{figure}

The goal of this section is to express Eq. \eqref{eq:BF_final} in the general 3D case, 
while the concepts discussed using the Bar-and-Flywheel models remain the same. 
Table \ref{table:task_space} will be important in translating this structure into 3D. 
Fig. \ref{fig:frames_def} shows the relevant frames, corresponding conceptually to Fig. \ref{fig:bar_and_flywheel_4}.

We begin by translating Eq. \eqref{eq:BF_HCoM_approx1} into 3D.  The 3D angular momentum is
\begin{equation} \label{eq:H_com_real}
H_{CoM} = M_B \omega_B + M_q \dot{q} = M_B [\omega_B + A \dot{q}],
\end{equation}
where $M_B$ and $M_q$ are base and joint space centroidal momentum matrices
\cite{orin2008centroidal},
$A \triangleq M_B^{-1} M_q$ is the local connection and a function of $q$, 
$\omega_B$ is the angular velocity of the base relative to the world expressed in the base frame, and $\dot{q}$ are the joint velocities.
For the WBO, we have an approximated angular momentum
\begin{equation} \label{eq:H_com_ACoM}
\tilde{H}_{CoM} = M_B [\omega_B + \tilde{A} \dot{q}] ,
\end{equation}
where $\tilde{A}$ will be the approximated local connection as discussed in arriving at Eq. \eqref{eq:BF_final}.
In 3D, trying to minimize the differences between these two representations (Eqs. \eqref{eq:H_com_real} and \eqref{eq:H_com_ACoM}) means making:  
\begin{equation} \label{eq:AqAq}
A\dot{q} \approx \tilde{A}\dot{q}. 
\end{equation}
We will keep $\dot{q}$ on both sides of the approximation until we sort out the 3D rotation representation for WBO in Section \ref{sec:parameterization}.
We note that $A\dot{q}$ on the left hand side of Eq. \eqref{eq:AqAq} is the \emph{relative angular velocity} of the system \cite{miyata2019walking}.
On the other hand, $\Omega_{Wbo} \triangleq  \tilde{A}\dot{q} $ is the angular velocity of the WBO frame relative to the base, expressed with respect to the base frame.

Paralleling the 2D example above (see Table \ref{table:task_space}), we would like to find a function for $Q \triangleq Q_{B,Wbo}$.
This represents the WBO frame orientation relative to the base.
Thus, $\Omega_{Wbo}$ can be expressed from the quaternion rate $\dot{Q}$ using
\vspace{-0mm} 
\begin{equation} \label{eq:Eq}
  \Omega_{Wbo} = 2 R_Q E_Q \dot{Q}
\end{equation}  
where $R_Q$ is the rotation matrix representation of $Q$, and the matrix $2 \cdot E_Q$ maps a quaternion rate to an angular velocity (details are omitted here for brevity; see \cite{wie1985quaternion}).

\begin{table}
\centering
\begin{tabular}{ || c || c | c  || } 
 \hline
  & Bar-Flywheel & in 3D \\ 
 \hline
 \hline
 joint configuration & $[d, \phi]$ & $q$ \\ 
 \hline
 base orient. r.t. world & $\theta$ & $Q_{W,B}$\\ 
 \hline
 WBO r.t. base & $\tilde{\psi}_{rel}$ & $Q \triangleq Q_{B,Wbo}$ \\ 
 \hline
 WBO r.t. world & $\tilde{\Psi}_{Wbo}$ & $Q_{W,Wbo}$ \\ 
 \hline
\end{tabular}
\vspace{1mm}

\ \ \ \ r.t. = relative to \ \ \ \ Q = Quaternion
\vspace{-2mm}
\caption{\vspace{-1mm}Notation definitions and correspondences}
\vspace{-2mm}
\label{table:task_space}
\end{table}

\begin{algorithm}[t!]
\caption{WBO optimization}
\begin{algorithmic}[1]\label{alg:acom_solver}
\renewcommand{\algorithmicrequire}{\textbf{Input:}}
\renewcommand{\algorithmicensure}{\textbf{Output:}}
\REQUIRE  $N$ random joint configurations $q_i, \ i = 1, ..., N$
\ENSURE  $\Theta^*$
\STATE $\Theta \leftarrow 0$ (initialize to constant identity rotation)
\REPEAT
\STATE Substitute $Q(q_i;\Theta)$ into $T_Q$  in Eq. \eqref{eq:full_opt_problem_param} for $ i = 1, ..., N$
\STATE $\Theta \leftarrow$ Solve Eq. \eqref{eq:full_opt_problem_param} with given $T_Q$
\UNTIL convergence
\RETURN $\Theta$
\end{algorithmic}
\end{algorithm}

\subsection{Parameterization and Optimization Algorithm} \label{sec:parameterization}

Noting that $Q$ has two portions $Q = [Q_s; \, Q_{x,y,z}]$,
we now parameterize $Q_{x,y,z}$ by a vector of basis functions $\lambda(q)$ with dimension $n_\lambda$:
\vspace{-1mm} 
\begin{equation}\label{eq:Q_and_param}
Q_{x,y,z} (q;\Theta) \, = \, \Theta \lambda(q),
\end{equation}
where $\Theta \in \mathbb{R}^{3 \times n_\lambda}$ is a coefficient matrix.
$Q_s$ can be recovered from the unit norm constraint $\| Q \|_2^2 = 1.$
Similarly, we take the time derivatives of $Q_{x,y,z}$ and recover $\dot{Q}_s$ from $\frac{d}{dt} \| Q \|_2^2 = 0$.
These algebraic manipulations will lead to a final form:
\begin{equation}\label{eq:Aq_final}
  \tilde{A} \dot{q} = T_Q \Theta J_{\lambda} \dot{q}
\end{equation}
where $T_Q$ and $J_{\lambda}$ are functions of $q$ and are respectively defined as
\begin{equation*}
\begin{alignedat}{2} 
& 
T_Q
\triangleq
2 R_Q E_Q
\left[
\begin{array}{@{}c@{}}
  -Q_s^{-1}Q_{x,y,z}^T\\
  I_{3 \times 3}\\
\end{array}
\right] 
\in \mathbb{R}^{3 \times  3}
\text{, and}\\
&
J_{\lambda} \triangleq \frac{\partial{\lambda(q)}}{\partial{q}}
\in \mathbb{R}^{n_\lambda \times  n_q}
.
\end{alignedat}
\end{equation*}
Given Eq. \eqref{eq:AqAq}, our objective is to minimize the difference between $A$ and $\tilde{A}$, both of which are functions of $q$. 
Since minimizing the difference over an infinite number of $q$ in a region of joint space is often intractable, 
we pre-select $N$ number of random configurations (uniformly distributed in the robot's operating joint space) to simplify the problem:
\begin{equation}\label{eq:full_opt_problem_param}
  \underset{\Theta}{\text{min}} \ 
  \frac{1}{N}
  \sum_{i=1}^{N} \| A_i - T_{Q_i} \Theta J_{\lambda i} \|^2_F 
\end{equation}
where $\| \cdot \|_F$ is the Frobenius norm.
Given our choice in Eq. \eqref{eq:Q_and_param}, 
we note that $A$ and $J_\lambda$ are independent of $\Theta$, while $T_Q$ is nonlinear in $\Theta$.

One can solve Eq. \eqref{eq:full_opt_problem_param} with many nonlinear solvers. 
In practice, we found that our simple algorithm in Alg. \ref{alg:acom_solver} works.
The algorithm exploits the structure of the cost function by identifying that Eq. \eqref{eq:full_opt_problem_param} is a least squares problem if $T_Q$ is given. 
In each iteration, we first substitute the current solution $\Theta$ into $T_Q$ to turn \eqref{eq:full_opt_problem_param} into a least squares problem\footnote[4]{
The Kronecker product identity
$vec(XYZ) = Z^T \otimes X \: vec(Y)$
is useful in vectorizing the matrix $\Theta$ in preparation for solving the least squares.
}, and then solve the problem to get a new solution $\Theta$. 
We repeat the above steps until $\Theta$ converges.  
This algorithm is similar to the Gauss-Newton method, differing in that it avoids linearizing the objective function at the solution in each iteration.

\begin{figure}[t!]
\centering

\subfigure[Angular velocities of one step of Nadia robot walking on flat ground. The solid and dashed lines are $A\dot{q}$ and $\tilde{A}\dot{q}$ in Eq. \eqref{eq:AqAq}, respectively.
The spikes around 0.6 seconds are from the swing foot impact event and the feedback reaction of the walking controller.]{\includegraphics[width=0.45\textwidth]{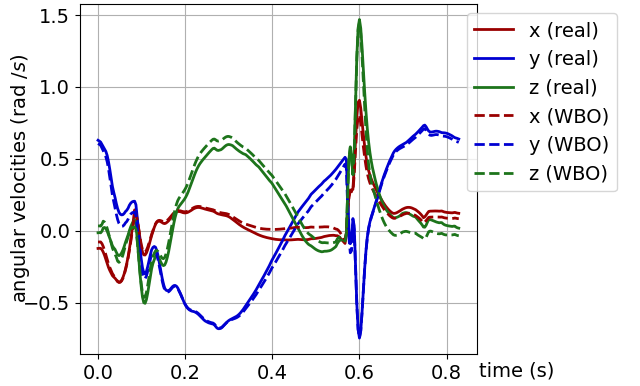}
\label{fig:real_vs_acom__angular_vel}}

\vspace{1mm}

\subfigure[The solid lines are the real CAM $H_{CoM}$ in Eq. \eqref{eq:H_com_real}, and the dashed lines are the approximated CAM by the WBO $\tilde{H}_{CoM}$ in Eq. \eqref{eq:H_com_ACoM}.]{\includegraphics[width=0.45\textwidth]{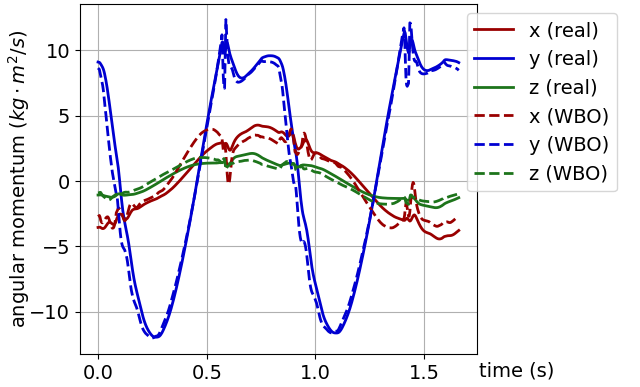}
\label{fig:real_vs_acom__CAM}}
\caption{Comparisons between real and approximated quantities. The data is from a simulation where Nadia walked in a straight line at $0.6$ m/s.}

\label{fig:real_vs_acom}
\end{figure}

\subsection{WBO Optimization and Result} \label{sec:acom_opt_result}

We optimized for an WBO function for Nadia (Fig. \ref{fig:frames_def}) using Alg. \ref{alg:acom_solver}.
Nadia is a humanoid robot with 31 degrees of freedom (DoF) -- 6DoF legs, 7DoF arms, 1DoF grippers and a 3DoF spine. 
We randomly select 1000 configuration pairs mirrored about the sagittal plane (so $N=2000$).
We also keep the gripper, wrist and ankle joints at neutral positions, because their contribution to the CAM is relatively small. 
This reduces the configuration space ($q$) to 19 dimensions.
The basis functions are monomials in terms $q$, with all possible monomials up to 3rd order being used (e.g. $q_i$, $q^2_i$, $q_iq_j$, $q^3_i$, $q^2_iq_j$, ...), producing 1539 basis functions. 
The optimization converged smoothly in about 2 minutes or 10 iterations. After the optimization, we dropped terms with coefficients (in $\Theta$) smaller than 1e-8.

In addition to Nadia, we also optimized a WBO function for Cassie running (Fig. \ref{fig:hardware_side_by_side_comparison}).
Cassie has 16 joints.
We ignored the toe and ankle spring joints during optimization, reducing the configuration space to 12 dimensions.
The optimization converged within 10 seconds and 7 iterations.

By comparing actual (measured) and approximated quantities, we can evaluate our WBO approximation at different signal scales.
For Nadia walking, Fig. \ref{fig:real_vs_acom__angular_vel} plots both sides of Eq. \eqref{eq:AqAq}.
Average angular velocity errors for each axis are about $[0.034, 0.035, 0.061]$ rad/s.
Fig. \ref{fig:real_vs_acom__CAM} plots Eqs. \eqref{eq:H_com_real} and \eqref{eq:H_com_ACoM}. 
We see that $H_{CoM}$ and $\tilde{H}_{CoM}$, which are larger signals dominated by base motion, are relatively close: 
average errors for each axis are about [0.74, 0.84, 0.32] $\text{kg} \cdot \text{m}^2 / \text{s}$.
Thus, WBO reflects the actual $H_{CoM}$ in a meaningful way.

\section{Walking example} \label{sec:walking_example}

In this section, we design a walking controller for Nadia using the WBO derived in Section \ref{sec:acom_opt_result}, and show that a WBO reference tracking can induce natural upper body motions during walking (Fig. \ref{fig:hardware_side_by_side_comparison}).

\subsection{Controller}

\begin{figure}[!t]
\centering
\subfigure[Controller with constant desired joint positions of the upper body.]{\includegraphics[width=0.49\textwidth]{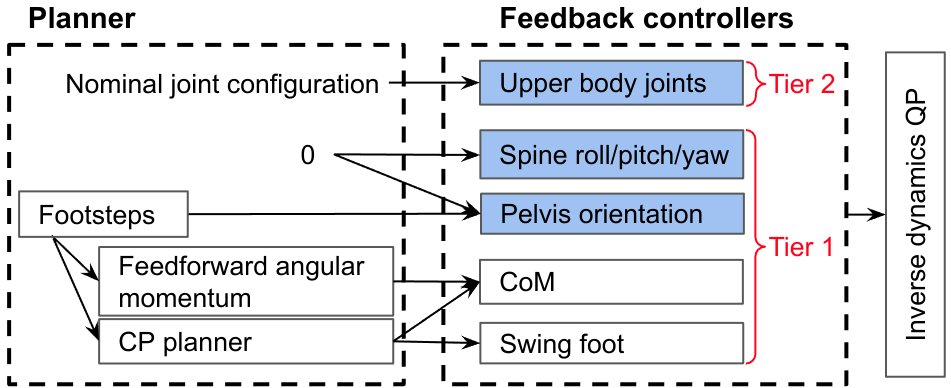}
\label{fig:controller_diagram_default}}

\vspace{2mm}

\subfigure[Controller with WBO-induced upper body motions.]{\includegraphics[width=0.49\textwidth]{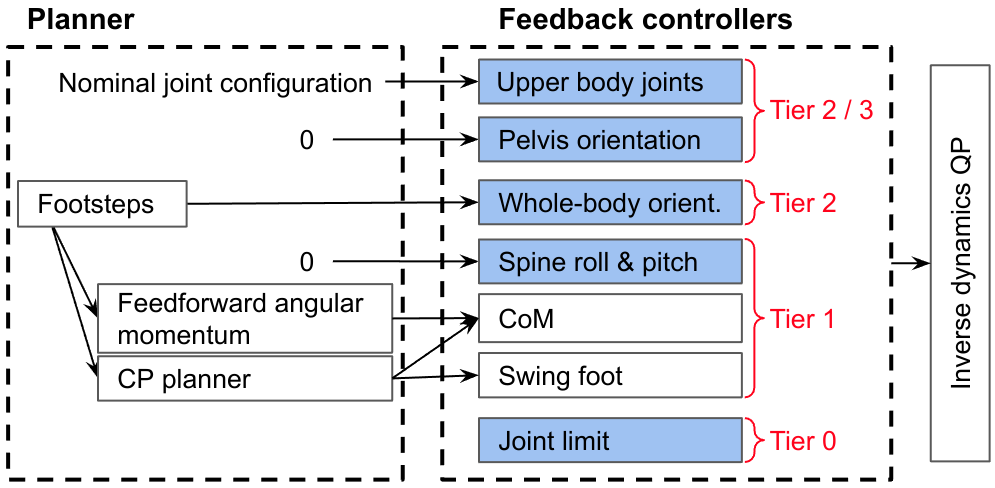}
\label{fig:controller_diagram_acom_ideal}}
\caption{Controller diagrams. 
     Each diagram is composed of a high-level planner, low-level feedback controllers and an inverse-dynamics quadratic program (QP).
     Blue color highlights the difference between the two controllers. 
     Red color is used for indicating the task priorities. 
     Tier 0 is implemented as a constraint in the QP, while other Tiers are implemented via cost functions in the QP.
     Additionally, Tier $n$ has higher priority than Tier $n+1$ for $n>0$.
     We use the null-space projection technique to prioritize tasks \cite{hutter2013hybrid}.
     The zeros and nominal joint configuration in the planner are constant trajectory sources.
         }
\label{fig:control_diagrams}
\end{figure}

Fig. \ref{fig:controller_diagram_acom_ideal} shows our WBO controller, 
while Fig. \ref{fig:controller_diagram_default} shows the baseline controller which fixes the desired joint positions for the upper body.
Each controller has a planner that generates desired trajectories.
These are then converted into acceleration commands by the feedback controllers shown in the diagrams.
With the acceleration commands, we use an inverse dynamics whole body controller (the rightmost block in each diagram in Fig. \ref{fig:control_diagrams}) to get the desired actuator commands for the robot \cite{Koolen15}.
We can roughly separate the controller into leg and upper body parts.
The leg part handles tracking the desired path and heading of the robot, while the rest of the controller handles the upper body motion.

\subsubsection{Legs}
This part of the controller is the same between the baseline and WBO controllers.
We use Capture Point (CP) control for the locomotion task \cite{koolen2012capturability, 
seyde2018inclusion}.
The footsteps are generated given desired velocity commands from a higher level controller.
The planner outputs a reference Centroidal Moment Pivot trajectory that is converted into a linear momentum rate command in the feedback controller, which is sent to the inverse dynamics controller.

\subsubsection{Upper body}

Our short-term goal was getting more natural arm swing and spine yaw rotation by servoing just WBO yaw\footnote[5]{
In our experiments, we found that arm swing and spine yaw rotation were mostly induced by servoing the WBO yaw angle to zero. 
Additionally, Miyata et al. \cite{miyata2019walking} only used the yaw part of angular momentum to generate the arm swing.
}.  
In Fig. \ref{fig:controller_diagram_acom_ideal}, we servo the WBO yaw axis relative to the world frame, while both the pelvis and the upper body tasks reside in the null space of the WBO task.
In simulation, we achieved straight-line walking with this controller.
When moving to hardware, we temporarily focused on demonstrating arm swing and spine yaw rotation. To do so, we servoed the pelvis orientation relative to the world and regulated the WBO yaw angle relative to the pelvis to 0.
More complex motions have been left to future work, where we would like to take full advantage of our WBO.

We use a task hierarchy \cite{hutter2013hybrid} in our whole body controller, shown in Fig. \ref{fig:control_diagrams} as ``Tiers". 
In experiments, we noticed that the inverse dynamics QP solver would trade swing foot orientation tracking performance for WBO tracking performance.
This happened when the robot could not regulate WBO yaw to 0 with only the upper body.
Thus, in order to prevent the WBO task from impairing the leg tasks, we set the WBO task to a lower priority than the leg tasks.

Besides the above task objectives, we also add nominal joint configuration tracking to handle the system's redundancy.
This task can exist in the null space of the WBO task or at the same level as WBO.
The parameters for this upper body joint controller can be used to sculpt the desired motion.
For example, increasing the cost weight on the spine joint achieves more arm swing and less spine rotation.

\subsubsection{Joint limits}
The joint limit controller takes current joint positions and ranges of motion, and outputs limits on joint accelerations for the whole body controller.
These limits are used for self-collision avoidance and aesthetics.
Because the legs on Nadia are much heavier than the arms, when regulating the WBO yaw to 0, the robot can generate excessive arm swing or spine rotation. Thus, self-collision avoidance helps contain these motions, and therefore affects WBO tracking and overall appearance.
We note that conventional momentum approaches \cite{erez2012trajectory, miyata2019walking} would also exhibit this same behavior on Nadia.

\begin{figure}[!t]
\centering
\subfigure[Simulation (average walking speed $\approx 0.6$ m/s)]{\includegraphics[width=0.4\textwidth]{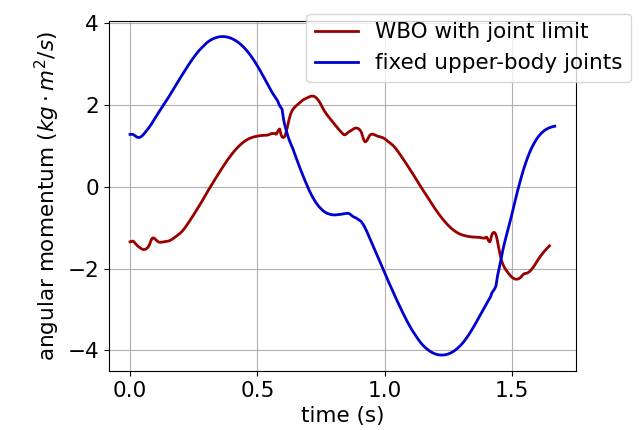}
\label{fig:CAM_experiment_simulation}}

\subfigure[Hardware (average walking speed $\approx 0.37$ m/s)]{\includegraphics[width=0.4\textwidth]{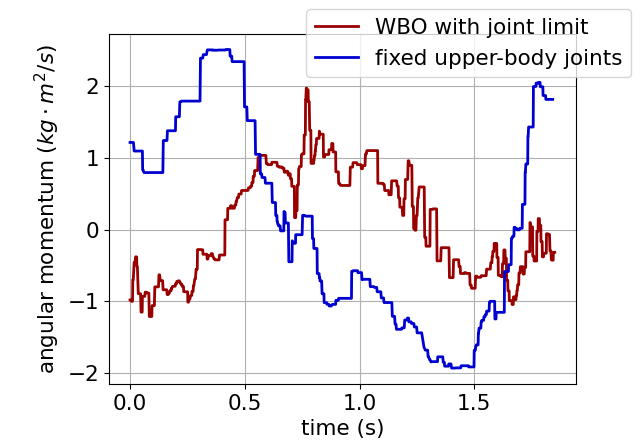}
\label{fig:CAM_experiment_hardware}}
\caption{The CAM about the z axis when Nadia walked in a straight line. 
We note that there were issues with Nadia's leg actuator at the time of hardware experiment and the update rate of the control loop was not fast. 
These issues partially caused the non-smoothness in the hardware plot. 
}
\vspace{0mm}
\label{fig:CAM_experiment}
\end{figure}

\subsection{Experiment Result}

In both simulation and hardware experiments, we saw natural upper body motion induced by tracking a constant WBO.
Additionally, although the controller for the upper body motion was designed for straight-line walking, 
we found that, in simulation, the robot was also able to walk forward, backward, sideways, and turn.
The video clips can be found in the accompanying video for this paper. 

Fig. \ref{fig:CAM_experiment} shows the z-component of CAM of straight-line walking for both the WBO and the baseline controller.
We see that the angular momentum profiles look similar between the simulation and hardware, 
and that the CAM is 50\% smaller when the desired WBO yaw is set to 0.
Joint limit constraints prevent the CAM from tracking closer to zero.
Additionally, we observed in simulation that the WBO controller reduces foot yaw moment against the ground.
These are some of the advantages of using the upper body to counter moments generated by the legs during walking \cite{miyata2019walking}.

The conventional approach to generating natural upper body motions is directly minimizing the CAM \cite{erez2012trajectory, miyata2019walking}. 
The downside of this approach is that the CAM controller is a feedback controller based on mutually constrained velocities rather than positions.  
Thus the upper body configuration could gradually drift away from a neutral target unless care is taken.
To address this, a competing control objective is typically introduced that servos the robot, or some specific link on the robot, back to a desired orientation relative to the world. 
In contrast, a control law based on the WBO provides a single desired orientation for the entire robot, and thus need not employ competing objectives.

\section{Running example} \label{sec:running_example}

\begin{figure}
 \centering
 \includegraphics[width=0.9\linewidth]{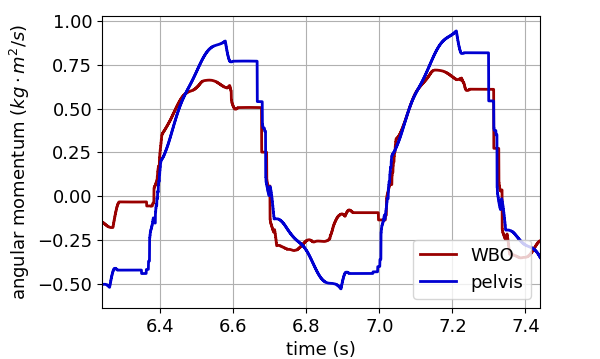}
 \caption{The CAM about the z axis when Cassie runs at 2.7 m/s and follows a desired yaw trajectory that goes from 0 to $\pi/2$ rad in 10 seconds.
}
 \label{fig:cassie_experiment}
\end{figure}

Besides the walking example, we also want to test our WBO on a slightly more agile motion.
For this, we implemented two running controllers on Cassie.

\subsection{Controller}
The baseline running controller uses a finite state machine with four states -- left stance, left flight, right stance and right flight.
The state transitions are triggered by foot touch-down and lift-off events.
In the left/right stance state, the stance leg behaves like a vertical virtual spring, the pelvis pitch and roll angles are regulated to 0, and pelvis yaw follows a desired trajectory. 
The swing leg uses a Raibert-style control law \cite{raibert1986legged}, while the leg length is determined by the nominal leg length at touchdown.
In the flight state, the controller continues to track the desired orientation of the pelvis and the desired positions of the leading swing foot. The desired pelvis orientation is relative to the world frame.

The second (preliminary) running controller is the same as the baseline controller, except that we replace the pelvis yaw with WBO yaw.

\subsection{Experiment Result}
In our experiments, Cassie is commanded to run at 2.7 m/s in the Drake simulator \cite{Drake2016}.
We also set a desired yaw trajectory, which goes from 0 to $\pi/2$ rad linearly in 10 seconds.
The baseline controller tracks this desired yaw with the pelvis, while the WBO controller tracks it with WBO.
Fig. \ref{fig:cassie_experiment} shows the CAM of Cassie. We can see that the momentum oscillates less with the WBO controller (more than a 26\% reduction).
This reduction is due to the WBO representing the orientation of the entire system, and the total momentum is approximated by its time derivative via Eq. \eqref{eq:H_com_ACoM}. 
Also, the WBO controller is able to adjust the desired pelvis orientation when the legs move.
In contrast, the baseline controller considers the pelvis motion only and ignores the contributions from the legs.
One could, of course, regulate the CAM while tracking the desired orientation of the pelvis, but these two objectives could conflict since the pelvis alone does not represent the entire system well.
The advantage of the WBO approach here is the consistency between the orientation-tracking and momentum objectives.

\section{Conclusion and future work}\label{sec:conclusion}
\vspace{-1mm}
We introduced the integrable whole-body orientation (WBO) with simple examples and clear problem motivation, 
so it is more accessible to a general robotic audience.
A formulation of the WBO problem was provided, including an algorithm that solves the problem quickly by exploiting its structure.
WBO functions were synthesized offline for the Nadia and Cassie robots, and were then used to induce arm swing and spine yaw rotation in a walking example and to turn the robot's global orientation in a running example. 

The WBO enables us to servo the orientation of the entire system.
Thus, it can free up the base link (e.g. pelvis) to achieve high-level goals such as natural walking with natural pelvis motion and stepping up/down terrain.
In this paper, we mostly demonstrated more natural arm swing and spine rotation.
Future work will utilize the WBO to achieve more complex behaviors, such as whole-body natural walking.
Another area of future research involves incorporating high-level planning for the WBO trajectories (e.g. using the SRB model in planning), which could potentially enable more agile motions for the robots.
Lastly, this paper does not explore the impact of WBO on system stability, leaving it for further investigation.
However,  a prior study \cite{posa2017balancing} found larger regions of attraction for balance and step recovery by moving from a point-mass to SRB model. 
We believe substituting WBO for the SRB model could also improve stability for the above systems.

\section{Acknowledgment}
\vspace{-2mm}
We thank Sylvain Bertrand, Brandon Shrewsbury, Evan Yu, James Foster and Stephen McCrory for software setup and instruction; 
Stefan Fasano and Joseph Godwin for helping with the hardware experiment;
Brian Acosta for generating Cassie data for WBO learning;
and William Yang for providing the baseline Cassie running controller to use as an example.
This work was supported by DAC Cooperative Agreement W911NF2120241, ONR Grant No. N00014-22-1-2593, and ONR Contract No. N00014-19-1-2023.
Toyota Research Institute also provided funds to support this work.
\vspace{-4mm}

\bibliographystyle{ieeetr}
\bibliography{library,yuming_library}

\balance

\end{document}